\documentclass[10pt,twocolumn,letterpaper]{article}

\usepackage[pagenumbers]{cvpr} 

\usepackage{pgf-pie} 
\usepackage{multirow}
\usepackage{pifont}
\newcommand{\cmark}{\textcolor{green!70!black}{\ding{51}}}
\newcommand{\xmark}{\textcolor{red}{\ding{55}}}
\usepackage{xcolor}
\usepackage{tikz}

\newcommand{\radarplotfour}[6]{
\pgfmathsetmacro\rA{#3/1.5}
\pgfmathsetmacro\rB{#4/1.5}
\pgfmathsetmacro\rC{#5/1.5}
\pgfmathsetmacro\rD{#6/1.5}
\filldraw[fill=#1, draw=#2, thick, opacity=0.55]
({cos(90)*\rA},{sin(90)*\rA}) --
({cos(180)*\rB},{sin(180)*\rB}) --
({cos(270)*\rC},{sin(270)*\rC}) --
({cos(0)*\rD},{sin(0)*\rD}) -- cycle;
}

\usepackage{graphicx}
\usepackage{booktabs}
\usepackage{xcolor}
\usepackage{array}
\usepackage{ragged2e}
\usepackage{caption}

\definecolor{cvprblue}{rgb}{0.21,0.49,0.74}

\usepackage[pagebackref,breaklinks,colorlinks,allcolors=cvprblue]{hyperref}

\def\paperID{10} 
\def\confName{CVPR}
\def\confYear{2026}

\title{LLaVA-LE: Large Language-and-Vision Assistant for Lunar Exploration}

\author{
Gokce Inal$^*$\quad
Pouyan Navard$^{*\dagger}$\quad
Alper Yilmaz\\[6pt]
The Ohio State University\\
Columbus, OH, USA\\[4pt]
{\tt\small inal.5@osu.edu \quad bnd.pouyan@gmail.com \quad yilmaz.15@osu.edu}
}

\begin{document}
\maketitle

\renewcommand{\thefootnote}{\fnsymbol{footnote}}
\footnotetext[1]{Equal technical contribution.}
\footnotetext[2]{Alumnus of The Ohio State University.}

\renewcommand{\thefootnote}{\arabic{footnote}}
\setcounter{footnote}{0}

\begin{abstract}


Recent advances in multimodal vision–language models (VLMs) have enabled joint reasoning 
over visual and textual information, yet their application to planetary science remains largely 
unexplored. A key hindrance is the absence of large-scale datasets that pair real planetary imagery 
with detailed scientific descriptions. In this work, we introduce \textbf{LLaVA-LE} (Large 
Language-and-Vision Assistant for Lunar Exploration), a vision–language model specialized for 
lunar surface and subsurface characterization. To enable this capability, we curate a new 
large-scale multimodal lunar dataset, \textbf{LUCID} (\textbf{LU}nar \textbf{C}aption 
\textbf{I}mage \textbf{D}ataset) consisting of \textbf{96k} high-resolution panchromatic images 
paired with detailed captions describing lunar terrain characteristics, and \textbf{81k} 
question-answer (QA) pairs derived from $\sim$20k images in the LUCID dataset. Leveraging this 
dataset, we fine-tune LLaVA using a two-stage training curriculum: (1) concept alignment for 
domain-specific terrain description, and (2) instruction-tuned visual question answering. We 
further design evaluation benchmarks spanning multiple levels of reasoning complexity relevant 
to lunar terrain analysis. Evaluated against GPT and Gemini judges, LLaVA-LE achieves a 
\textbf{3.3$\times$} overall performance gain over Base LLaVA and \textbf{2.1$\times$} over 
our Stage 1 model, with a reasoning score of \textbf{1.070} — \textit{exceeding the judge's 
own reference score} — highlighting the effectiveness of domain-specific multimodal data and 
instruction tuning to advance VLMs in planetary exploration.%
\footnote{Code: \url{https://github.com/OSUPCVLab/LLaVA-LE}}

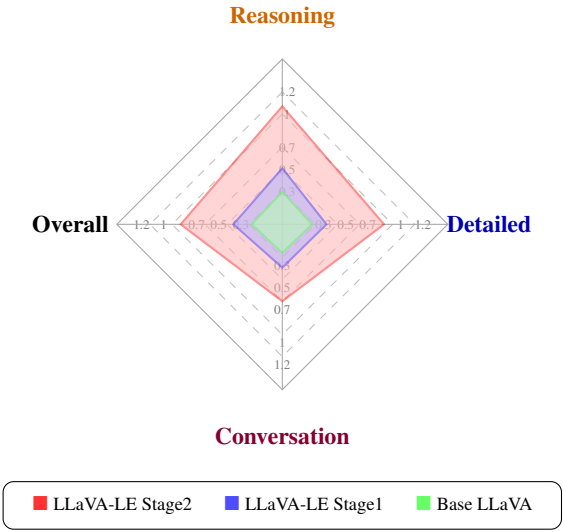
\begin{figure}[t]
\centering
\begin{tikzpicture}[scale=2.2]

\foreach \val in {0.3, 0.5, 0.7, 1, 1.2}{
    \pgfmathsetmacro\r{\val/1.5}

    \draw[dashed, gray!50]
    ({cos(90)*\r},{sin(90)*\r}) --
    ({cos(180)*\r},{sin(180)*\r}) --
    ({cos(270)*\r},{sin(270)*\r}) --
    ({cos(0)*\r},{sin(0)*\r}) -- cycle;

    \node[font=\tiny, gray] at ({cos(90)*\r + 0.03},{sin(90)*\r}) {\val};
    \node[font=\tiny, gray] at ({cos(180)*\r - 0.05},{sin(180)*\r}) {\val};
    \node[font=\tiny, gray] at ({cos(270)*\r},{sin(270)*\r - 0.05}) {\val};
    \node[font=\tiny, gray] at ({cos(0)*\r + 0.05},{sin(0)*\r}) {\val};
}

\draw[gray!70]
({cos(90)},{sin(90)}) --
({cos(180)},{sin(180)}) --
({cos(270)},{sin(270)}) --
({cos(0)},{sin(0)}) -- cycle;

\foreach \ang in {0,90,180,270}
\draw[gray!70] (0,0) -- ({cos(\ang)},{sin(\ang)});

\node[font=\small\bfseries, orange!80!black] at ({cos(90)*1.25},{sin(90)*1.25}) {Reasoning};
\node[font=\small\bfseries]                  at ({cos(180)*1.28},{sin(180)*1.28}) {Overall};
\node[font=\small\bfseries, purple!70!black] at ({cos(270)*1.28},{sin(270)*1.28}) {Conversation};
\node[font=\small\bfseries, blue!70!black]   at ({cos(0)*1.25},{sin(0)*1.25}) {Detailed};



\radarplotfour{red!30}{red!80}{1.070}{0.921}{0.698}{0.922}
\radarplotfour{blue!30}{blue!70}{0.508}{0.443}{0.393}{0.400}
\radarplotfour{green!30}{green!60}{0.295}{0.278}{0.260}{0.270}

\node[draw, rounded corners, fill=white, font=\scriptsize, inner sep=5pt]
at (0,-1.7)
{
\begin{tabular}{lll}
\textcolor{red!80}{$\blacksquare$} LLaVA-LE Stage2 &
\textcolor{blue!70}{$\blacksquare$} LLaVA-LE Stage1 &
\textcolor{green!60}{$\blacksquare$} Base LLaVA
\end{tabular}
};

\end{tikzpicture}

\caption{\textbf{Category-wise Visual Question Answering Performance Relative to Judge Scores}. We evaluate Base LLaVA, LLaVA-LE Stage 1, and LLaVA-LE Stage 2 on a held-out evaluation set across three question categories: Detailed, Conversation, and Reasoning. Each axis shows the model's score normalized by the judge's score (Score / Judge Score), averaged across GPT-4 and Gemini judges. The Overall axis represents the mean across all three categories. Values greater than 1.0 indicate the model outperforms the judge's reference score.}

\label{fig:radar_avg_judges}

\end{figure}

\end{abstract}    
\section{Introduction}
\label{sec:intro}

\begin{figure*}[t]
\centering
\includegraphics[width=\linewidth]{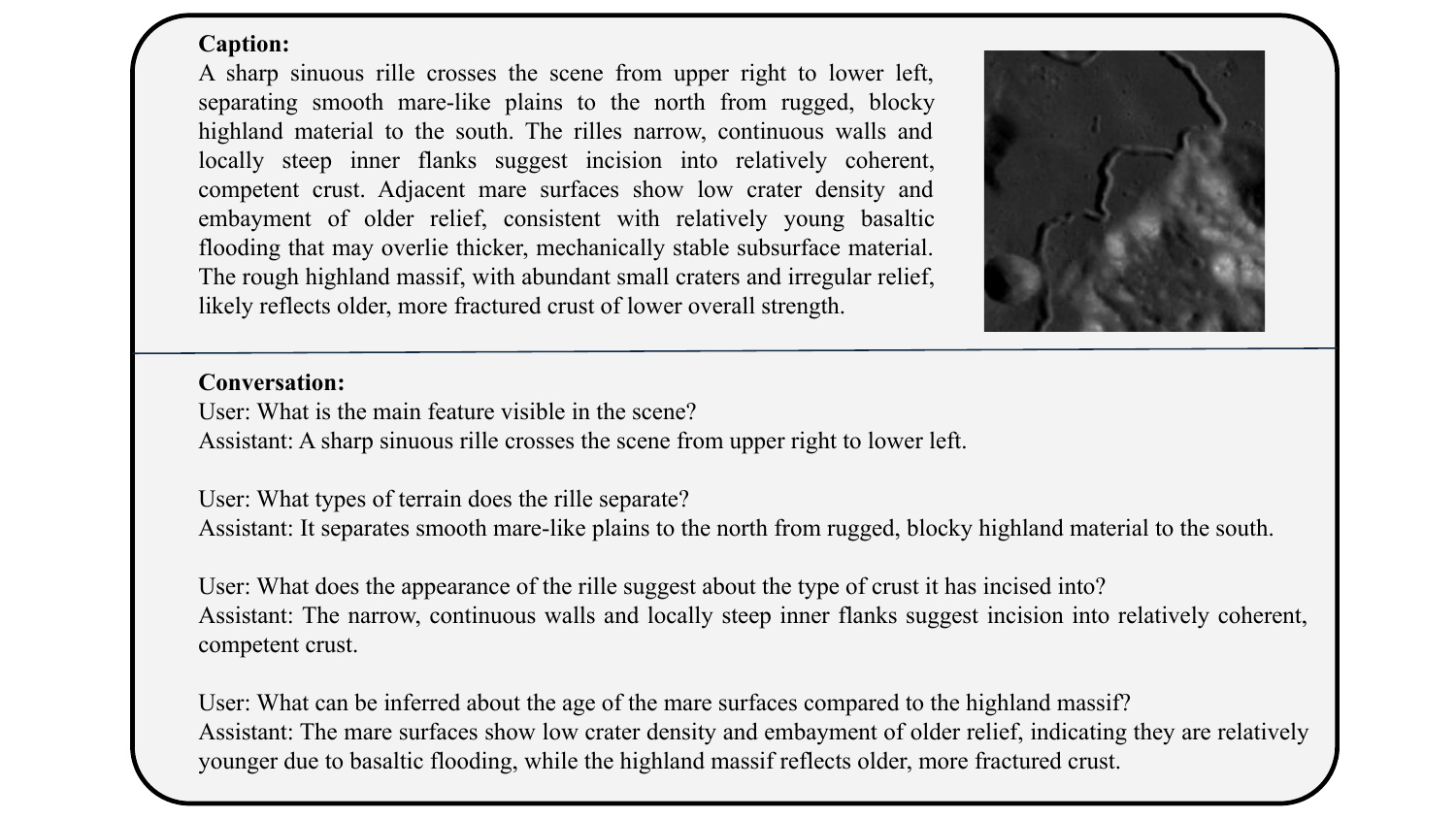}
\caption{\textbf{An Example of LUCID dataset}: The top shows the the panchromatic image with the associated caption generated using GPT, while the below shows question-answer pairs generated from that caption.}
\label{fig:instruction_example}
\end{figure*}

The wide availability of image-text datasets on the Internet has driven significant progress in vision-language modeling. Large-scale collections of natural images paired with descriptive captions provide an ideal foundation for generative pretraining, where models learn visual-linguistic associations in a self-supervised manner. This approach allows networks to build rich joint representations of vision and language without relying on labor intensive and costly manual labels. Using large-scale image-text pairs, models such as CLIP \cite{radfordclip}, BLIP \cite{li2022blip} and LLaVA \cite{liu2023llava} have demonstrated strong visual reasoning capabilities in general domain settings. Instruction-tuning has been shown to equip large vision-language models with the ability to generalize across tasks in a zero-shot manner, completing a wide range of vision-language objectives without task-specific supervision. This framework has opened up the way for developing general-purpose multimodal conversational systems that can describe, interpret, and reason about visual scenes in a human-like manner.

Despite the fact that these advances have transformed general-domain adaptation for various applications, they struggle to generalize to planetary exploration where there is a lack of large-scale, high-quality paired image-text corpora. Existing lunar datasets~\cite{liu2024lusnar, jiao2025luseg, martens2024synthetic} are uni-modal, small in scale, frequently contain mixtures of synthetic simulations and real observations, and often suffer from limited spatial resolution or mismatched distributions between modalities. Notably, the Space-LLaVA~\cite{foutter2024space} fined-tuned LlaVA model on synthetic multi-modal data for extraterrestrial applications. However, neither the data nor the codebase were publicly made available for Space-LLaVA at the time of submission of our paper. Such limitations make them unsuitable for training modern vision-language models motivating construction of a large scale, high quality training data corpus of real lunar imagery.

Planetary remote sensing is fundamentally different from natural image understanding: interpreting a lunar region requires reasoning across multiple physical modalities, not just visual appearance. Although the relationship between images and text is largely self-contained in natural vision tasks, planetary remote sensing is inherently multimodal. In natural images, visual patterns reflect underlying physical and geological processes rather than the semantic object categories typically present in natural imagery. A single optical image often provides an incomplete representation of the geological structure of a region, revealing only surface reflectance patterns. Complementary modalities, such as gravity anomaly maps or terrain slope models, provide additional evidence on subsurface mass distributions, regolith structure, and morphological evolution. These modalities capture different physical aspects of the same region and must be interpreted jointly to produce meaningful geological insights. As a result, planetary analysis fundamentally requires reasoning across multiple co-registered data layers that together characterize both surface morphology and subsurface structure.

Despite the scientific importance of such multimodal reasoning, there currently exists no vision-language assistant trained on large-scale real planetary datasets capable of performing high-level geological interpretation from multimodal remote sensing observations. Prior efforts in multimodal planetary analysis rely primarily on small curated datasets or synthetic simulations, limiting their applicability to real-world exploration scenarios. To address this gap, we introduce \textbf{L}arge \textbf{L}anguage \textbf{A}nd \textbf{V}ision \textbf{A}ssistant for \textbf{L}unar \textbf{E}xploration (LLaVA-LE), a vision-language model designed for multimodal geological understanding and reasoning over lunar remote sensing data.

Our approach builds upon multi-modal reasoning capability of the LLaVA framework and adapts it to the planetary science domain through the creation of a large-scale multimodal lunar dataset derived from real NASA mission observations. Specifically, we curate co-registered data products from the Lunar Reconnaissance Orbiter Camera (LROC)~\cite{robinson2010lroc}, Gravity Recovery and Interior Laboratory (GRAIL)~\cite{zuber2013grail}, and Lunar Orbiter Laser Altimeter (LOLA)~\cite{smith2010lola} missions, representing high-resolution optical imagery, gravity anomaly measurements, and terrain slope maps. From these mission products, we constructed aligned multimodal image triplets that jointly capture surface morphology and subsurface gravitational structure. To provide rich semantic supervision, we generate detailed scientific descriptions using a structured prompting pipeline with GPT-5, producing captions that describe the geological context, terrain morphology, and inferred subsurface characteristics. These multimodal image–text pairs are subsequently transformed into instruction–response examples used to train a planetary vision-language assistant capable of describing lunar terrain, answering geological questions, and performing multimodal reasoning across planetary data layers. Our contributions are as follows:

\begin{itemize}

\item \textbf{LLaVA-LE}. We introduce a vision-language model specialized for planetary science, adapting the LLaVA framework to perform multimodal reasoning over  on \emph{real} lunar remote sensing data. Our training pipeline leverages a curriculum-style instruction generation process that enables the model to learn geological interpretation and scientific query answering from multimodal observations.

\item \textbf{LUCID}. We release the \textbf{first} real large-scale \emph{real} multimodal lunar dataset for vision-language learning. LUCID consists of \textbf{96k} panchromatic imagery paired with detailed scientific captions, as well as \textbf{81K} question-answer pairs as a VQA dataset. LUCID (LUnar Caption Image Dataset) enables scalable training and evaluation of multimodal models for lunar surface and subsurface analysis.

\item \textbf{Open Source}. To facilitate further research in AI-driven planetary exploration, we publicly release the LUCID dataset, the codebase, and LLaVA-LE model checkpoints.

\end{itemize}

\section{Related Work}
\label{sec:related_work}

\textbf{Visual Instruction Tuning.}
Visual instruction tuning aims to align vision-language models (VLMs) with natural language instructions using image-response pairs \cite{huang2025survey}. Although early work in large language models (LLMs) such as ChatGPT \cite{chatgpt}, GPT-4 \cite{peng2023instruction}, and open-source variants like Alpaca \cite{zhang2024alpaca}, Vicuna \cite{vicuna2023}, and Dolly \cite{dolly2023} established the efficacy of instruction following, LLaVA \cite{liu2023llava} pioneered this in the multimodal domain by connecting a CLIP \cite{radfordclip} visual encoder to an LLM. Subsequent architectures such as KOSMOS-1 \cite{huang2023kosmos1}, BLIP-2 \cite{blip2}, and Flamingo \cite{alayrac2022flamingo} explored various fusion strategies. Extensions such as MiniGPT-v2 \cite{chen2023minigptv2}, mPLUG-Owl2 \cite{ye2024mplug2}, Otter \cite{li2025otter}, StableLLaVA \cite{stablellava}, PandaGPT \cite{su2023pandagpt}, and LAMM \cite{yin2023lamm} further scaled these methods to multi-tasking and diverse modalities. Our work builds on the LLaVA framework, adapting its conversational capabilities to the specialized constraints of planetary imagery.

\textbf{Domain-Specific Visual Assistants.}
Specialized instruction tuning has seen significant success in technical fields, particularly medicine. Text-based clinical assistants \cite{li2023chatdoctor, han2023medalpaca, toma2023ccamel, xiong2023doctorglm, wang2023huatuo} paved the way for multimodal medical models. LLaVA-Med \cite{li2023llavamed} and related frameworks like OphGLM \cite{gao2023ophglm}, Qilin-Med-VL \cite{liu2023qilin}, Med-Flamingo \cite{moor2023medflamingo}, PCLMed \cite{yang2024pclmed}, MedBLIP \cite{chen2024medblip}, CheXagent \cite{chen2024chexagent}, PathAsst \cite{sun2024pathasst}, SkinGPT-4 \cite{zhou2023skingpt}, and SigPhi-Med \cite{zhou2025sigphi} showed that fine-tuning on domain-specific data (e.g., radiology, pathology) significantly outperforms general-purpose models.

Beyond medicine, instruction tuning has been applied to robotics such as PaLM-E \cite{driess2023palme},  EmbodiedGPT \cite{mu2023embodiedgpt}, visual programming \cite{suris2023vipergpt, gupta2023visprog}, and editing \cite{brooks2023instructpix2pix}. Although AlphaEarth \cite{alphaearth} indicates a growing interest in earth observation multimodal data, the application of these techniques to planetary science remains nascent. We address this gap by introducing a LLaVA-based approach tailored for lunar surface analysis.

\begin{figure}[t]
\centering
\includegraphics[width=\linewidth]{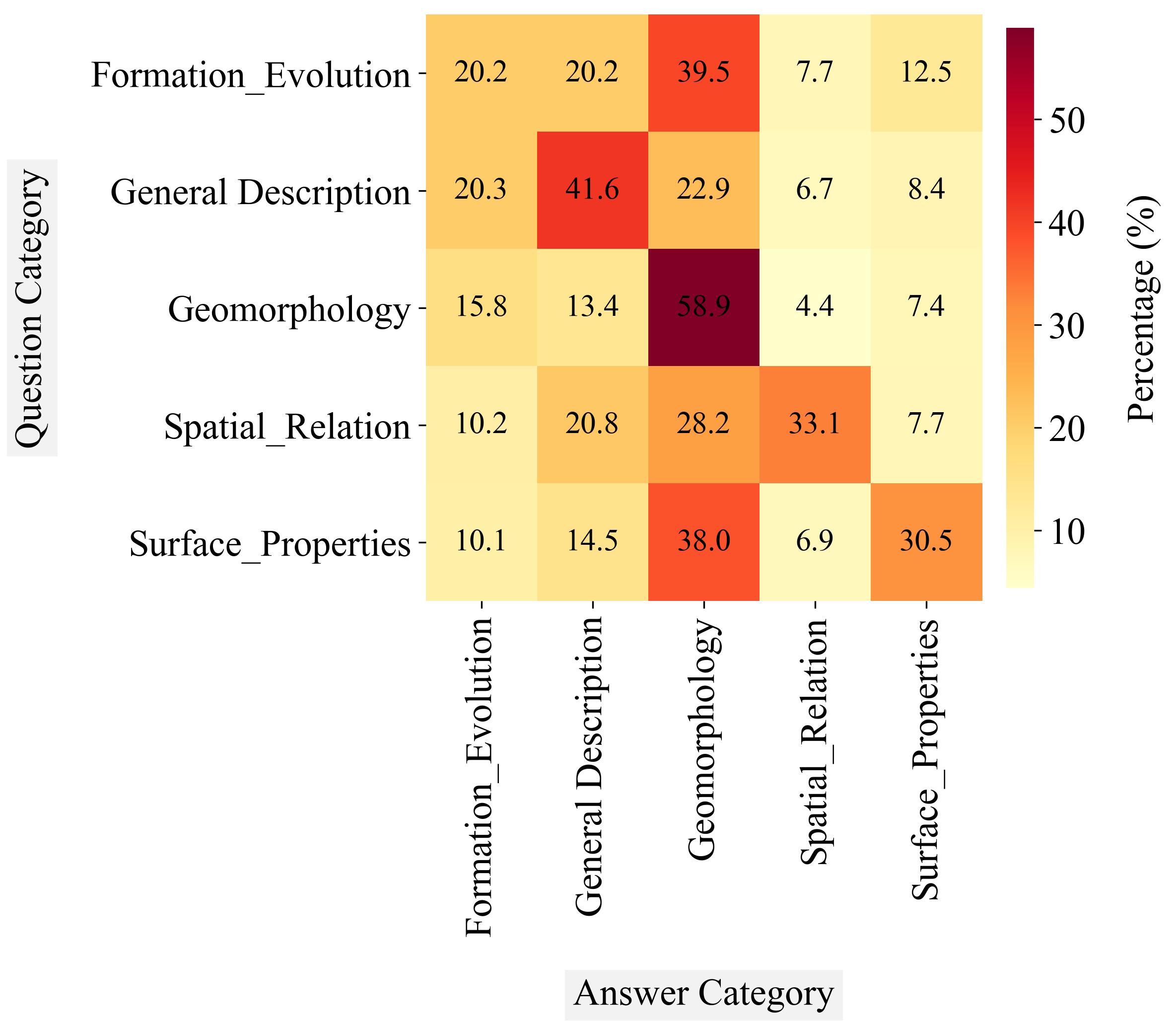}
\caption{\textbf{Distribution of LUCID VQA dataset across question-answer pairs categories.} This heatmap matrix shows how much each answer is contributing, in terms of percentage, to each one of the question categories across the entire LUCID visual question answering dataset. The heatmap is normalized by the row dimension and shows the reasoning capability captured in our dataset to answer each question.}
\label{fig:transitions}
\end{figure}

\begin{figure*}[t]
\centering
\includegraphics[
  width=\textwidth,
  trim=0cm 3.5cm 0cm 3.5cm,
  clip
]{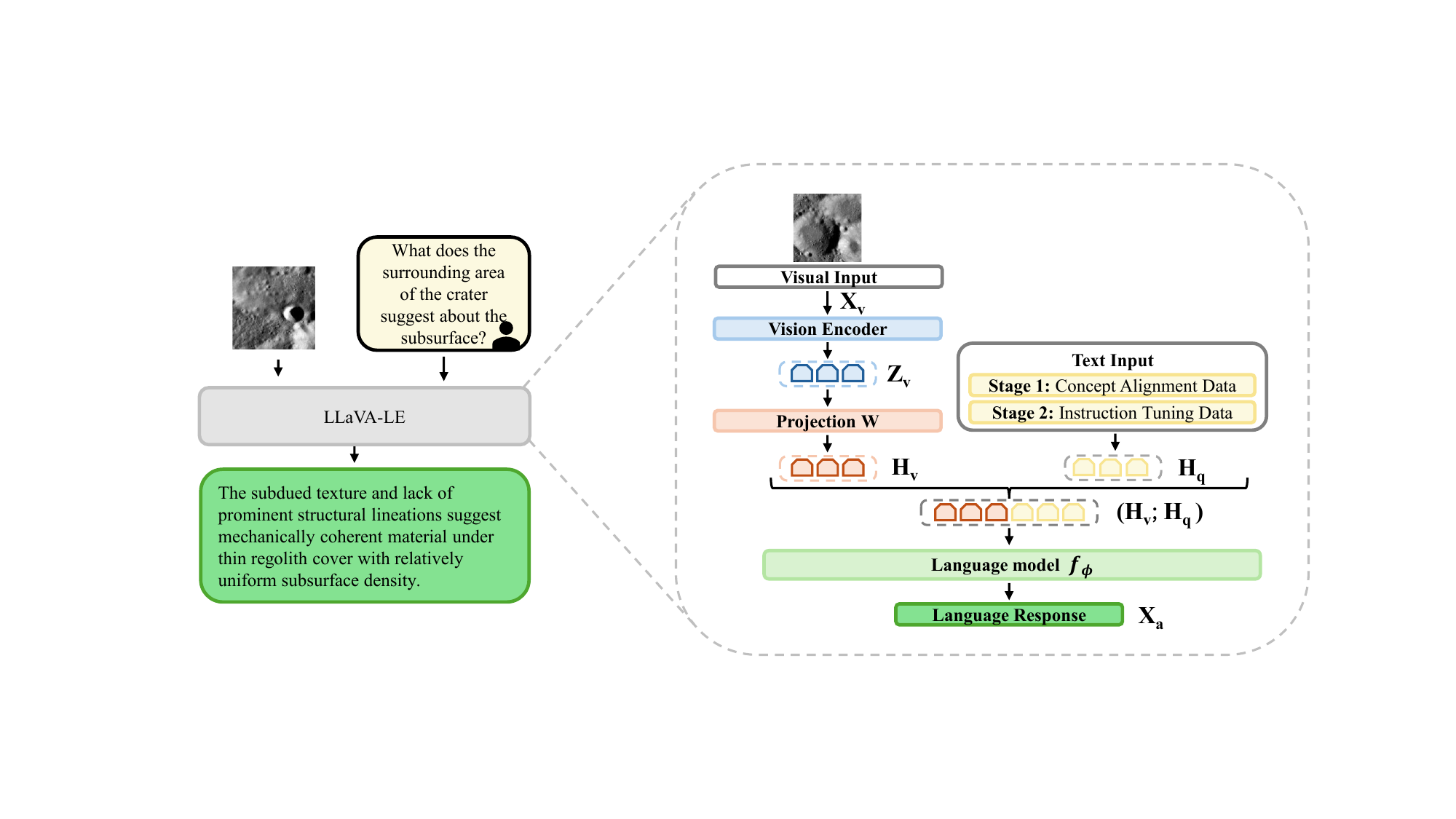}
\caption{\textbf{Overview of LLaVA-LE Training and Inference.} 
The right panel illustrates the two-stage training pipeline. Visual inputs $X_{\texttt{v}}$ 
are encoded by a frozen CLIP vision encoder $g$ to produce visual features $Z_v = g(X_{\texttt{v}})$, 
which are then projected into the language embedding space via a trainable projection layer, 
yielding $H_{\texttt{v}}$ to be concatenated with the language token embeddings $H_{\texttt{q}}$. 
The language model backbone $f_{\phi}$ autoregressively generates a response $X_{a}$ conditioned 
on the joint visual and language context. In \textbf{Stage 1 (Concept Alignment)}, caption 
supervision aligns lunar visual features with domain-specific geological language. In 
\textbf{Stage 2 (Instruction Tuning)}, the aligned model is further fine-tuned on multi-turn 
question-answer pairs to enable reasoning and conversational interaction. Throughout both stages, 
the CLIP vision encoder and the pretrained language model backbone $f_{\phi}$ remain frozen; 
only the projection layer and lightweight LoRA adaptation modules are updated. The left panel 
illustrates a representative inference example after Stage 2, where a user query about lunar 
surface observations is processed and the model generates a domain-grounded geological 
interpretation $X_{a}$.}
\label{fig:architecture}
\end{figure*}

\section{Method}

In this section, we describe the data and training procedure used to develop \textbf{LLaVA-LE}. We first introduce LUCID (\textbf{L}unar \textbf{C}aptioned \textbf{I}mage \textbf{D}ata) in section~\ref{data:caption_image_data}, a large-scale multimodal dataset designed to enable vision-language learning for lunar surface analysis. LUCID is constructed in two stages to support both planetary domain specific learning and instruction-following behavior. We then describe how Llava-LE is adapted from LLaVa using the LUCID dataset in section~\ref{model:llavale}.

\subsection{Lunar Captioned Image Data (LUCID)}
\label{data:caption_image_data}

To enable multimodal learning on real planetary observations, we constructed LUCID, a large-scale dataset designed for vision-language training on lunar surface analysis. LUCID contains \textbf{96K} samples derived from co-registered lunar remote sensing observations. The dataset is organized into two stages that mirror the training procedure for modern vision-language assistants. The first stage contains \textbf{76K} samples used for \textit{concept alignment}, where the model learns to associate visual patterns with geological descriptions. The second stage contains approximately \textbf{20K} samples used for \textit{instruction tuning}  for LLaVA-LE, where the caption data are transformed into question–answer interactions that simulate how users may query a lunar exploration assistant.

There is currently no publicly available multimodal dataset in planetary science that supports the training of vision-language models on \emph{real} planetary observations. \textbf{LUCID} addresses this gap by generating captions for high-resolution panchromatic imagery while leveraging complementary geophysical slope and gravity maps to ground the descriptions in scientifically meaningful observations. In the first stage, we generate image–caption pairs describing the morphology of the lunar surface across diverse terrains. In the second stage, these captions are converted into multi-round question–answer dialogs that emulate interactions between a user and a lunar exploration assistant. Figure~\ref{fig:instruction_example} shows an instance from our LUCID dataset. 

\subsubsection{Concept Alignment Data Generation}
\label{data:caption_alignment}

Decades of lunar remote-sensing research demonstrate that panchromatic images are one of the primary data sources for studying tectonics, volcanic history, and the evolution of stress fields on the Moon~\cite{zhang2024rilles, wang2024interpretation, moore1980lunar}. However, accurate geological interpretation typically requires analyzing panchromatic images in conjunction with complementary geophysical measurements. The surface morphology observed in panchromatic images is closely related to the underlying physical properties of the crust, including variations in slope, crustal thickness, and mass distribution measured by gravity missions~\cite{lro, smith2010lunar, nahm2023tectonics, scott1974geologic, shevchenko2012modern, zhang2024rilles}. Features such as the texture of the plains, the sharpness of topographic boundaries, the morphology of the crater rim, and the organization of structural lineaments are often manifestations of geological processes that also produce characteristic patterns in slope and gravity fields. Consequently, reliable interpretation of lunar surface and subsurface structure benefits from jointly analyzing panchromatic images with slope and gravity data, allowing surface observations to be grounded in the broader geophysical context.

Motivated by these well-established relationships between surface morphology and 
underlying geological structure, we first assemble a large set of lunar panchromatic 
image tiles from publicly available resources from the Lunar Reconnaissance Orbiter 
Camera mission \cite{robinson2010lroc}. Each image tile is partitioned into uniform 
spatial patches to create a consistent training corpus. The original tiles of 
$1504 \times 832$ px at a resolution of 125 m/px were each subdivided into 60 patches of $224 \times 224$ px, preserving the native resolution 
so that each patch spans approximately $28 \times 28$ km, corresponding to a 
surface coverage per-patch of ${\sim}784$ km$^2$. For every patch, we generate a 
corresponding scientific description by querying GPT-5.1 using 
a structured prompt designed to produce geophysically-grounded detailed captions of the observed terrain. 

This caption generation process is informed by the scientific relationships between surface morphology and geophysical structure. When generating captions, GPT-5.1 is instructed to focus on observable terrain attributes such as surface texture, crater morphology, structural lineaments, and shading patterns indicative of local topography. In addition to the panchromatic imagery, co-registered gravity anomaly maps and terrain slope measurements are provided as reference modalities to guide the interpretation of surface patterns during caption generation. Incorporating these complementary data sources encourages captions that reflect both visible surface features and inferred subsurface characteristics using slope and gravity. To the best of our knowledge LUCID is the first publicly available dataset for multi-modal lunar surface analysis.

These captions subsequently serve as the foundation for constructing the instruction-following dataset used in the second training stage.

\subsubsection{Instruction-Tuning Data}

To align the model with instruction-following behavior in the planetary science setting, we randomly selected $\sim$20K samples from our caption dataset (held out from the first-stage training described in Section~\ref{model:concpet_alignment_stage}) and prompted GPT-5.1 to generate 81K question–answer pairs, forming a VQA dataset derived from LUCID. This process transforms static image–caption pairs in Section~\ref{data:caption_alignment} into structured dialogues that resemble how a scientist or mission operator may interact with a lunar exploration assistant.

Given a panchromatic image $X_v$ and its corresponding caption $X_c$, we construct an instruction query $X_q$ that asks the model to analyze the terrain depicted in the image. In its simplest form, an instruction-following interaction can be expressed as:
\[
\begin{array}{c}
\text{Human:}\; X_q,\; X_v\; \langle \text{STOP} \rangle \\
\text{Assistant:}\; X_c\; \langle \text{STOP} \rangle
\end{array}
\]

This formulation enables the model to learn to produce descriptive geological explanations conditioned on both the image and the instruction prompt. An example of the generated instruction-following data is shown in Figure~\ref{fig:instruction_example}.

The generated questions-answers are designed to capture several core aspects of lunar terrain analysis, including surface morphology, geological processes that may explain the observed structures, indicators of relative surface age such as crater degradation and resurfacing, and localized spatial relationships between terrain features. For each caption, GPT-5.1 generates between three to five question–answer pairs that follow these guidelines, producing diverse conversational training examples grounded in the same underlying scene description.

Figure~\ref{fig:transitions} demonstrates the strong reasoning ability present in the question and answer pairs of the LUCID dataset across all five categories: Formation \& Evolution, General Description, Geomorphology, Spatial Relation and Surface Properties. To perform this analysis, we applied a lightweight NLP pipeline using spaCy to extract verbs and nouns from both question and answer turns, which were then classified into the five categories using a weighted lexicon-based scoring scheme. As shown in Figure~\ref{fig:transitions}, answering a question from any single category consistently draws upon reasoning across multiple categories, indicating that the dataset does not encourage narrow responses but instead promotes broad, multi-dimensional reasoning — a key property for training capable vision-language models.


\begin{figure*}[t]
\centering

\noindent\rule{\textwidth}{0.6pt}
\textbf{Visual Input: A Lunar Patch from LUCID Dataset}
\noindent\rule{\textwidth}{0.6pt}

\vspace{0.5em}

\begin{minipage}{0.32\textwidth}
    \centering
    \includegraphics[width=1.1in]{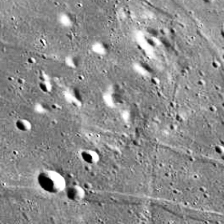}
\end{minipage}
\hfill
\begin{minipage}{0.64\textwidth}
\footnotesize
\textbf{Caption:} The dominant visible features are a cluster of small to medium impact craters scattered across a relatively smooth plains surface. Several craters in the lower part of the frame display sharp rims and well-defined interior walls, indicating locally steeper relief, whereas surrounding inter-crater areas appear comparatively flat. Subtle linear troughs and grooves trend diagonally across the scene, locally deflecting crater rims and ejecta. The combination of smooth plains and aligned structures may reflect a \textcolor{teal}{mechanically coherent}, \textcolor{teal}{relatively dense} substrate beneath a \textcolor{teal}{thin regolith cover}.
\end{minipage}

\vspace{0.8em}
\noindent\rule{\textwidth}{0.4pt}

\footnotesize
\textbf{User Question:} What does the crater distribution suggest about the underlying substrate?

\vspace{0.6em}
\noindent\rule{\textwidth}{0.4pt}

\footnotesize
\begin{tabular}{p{0.28\textwidth} p{0.68\textwidth}}
\textbf{Instruction Tuned LLaVA-LE} \cmark &
The crater distribution suggests a \textcolor{teal}{mechanically coherent}, \textcolor{teal}{relatively dense} substrate beneath a \textcolor{teal}{thin regolith cover.}\\
\end{tabular}

\vspace{0.5em}
\noindent\rule{\textwidth}{0.4pt}



\begin{tabular}{p{0.28\textwidth} p{0.68\textwidth}}
\textbf{Base LLaVA} \xmark &
The crater distribution on the moon's surface suggests that the underlying substrate is likely composed of a mix of hard and soft materials. The presence of small craters indicates that the surface is not entirely hard and rocky, as these craters have formed due to impacts from smaller objects. The larger crater in the middle of the image, which is more prominent and surrounded by smaller craters, implies that the substrate may have varying levels of hardness and resistance to impacts. This diversity in crater sizes and distribution could be due to the moon's geological history, which has resulted in a complex and varied surface composition.\\
\end{tabular}

\vspace{0.5em}
\noindent\rule{\textwidth}{0.4pt}
\begin{tabular}{p{0.28\textwidth} p{0.68\textwidth}}
\textbf{Chat GPT} &
The crater distribution, with sharp rims and well-defined interior walls, suggests that the underlying substrate is \textcolor{teal}{mechanically coherent and relatively dense}. This is indicated by the presence of smooth plains and the alignment of structures, which may reflect a solid substrate beneath a \textcolor{teal}{thin regolith} cover.
\\
\end{tabular}

\vspace{0.5em}
\noindent\rule{\textwidth}{0.4pt}
\begin{tabular}{p{0.28\textwidth} p{0.68\textwidth}}
\textbf{Gemini} &
The crater distribution, observed across a relatively smooth plains surface, contributes to the overall combination of features that may reflect a \textcolor{teal}{mechanically coherent}, \textcolor{teal}{relatively dense} substrate beneath a \textcolor{teal}{thin regolith cover}.
\\
\end{tabular}

\vspace{0.6em}
\noindent\rule{\textwidth}{0.6pt}

\caption{\textbf{Qualitative comparison of LLaVA-LE visual chat and reasoning capabilities in instruction-tuning tasks.} Comparison between the response of LLaVA-13B model and our LlaVA-LE model in during instruction-tuning stage evaluation.  The base LLaVA model often produces fluent but generic answers, occasionally lacking domain-specific terminology or geological interpretation. In some cases, the model hallucinates features that are not clearly implied by the scene or shifts toward broad discussions of lunar history rather than addressing the detail in the question. The Gemini and GPT are considered as the performance upper bound because the ground truth caption is fed into both of these models as context to generate the ground truth response.}
\label{fig:qualitative_examples}

\end{figure*}

\subsection{LLaVA-LE}
\label{model:llavale}
We initialize LLaVA-LE from a general-domain multimodal conversational model, LLaVA-v1.5-13B \cite{liu2023llava}, and fine-tune it on LUCID for lunar analysis. The method follows a two-stage training strategy designed to first build a basic multimodal understanding and then teach the model how to answer questions about lunar observations more specifically.

\subsubsection{Concept Alignment Stage}
\label{model:concpet_alignment_stage}
Stage 1 focuses on aligning lunar visual concepts with the language model representation space before introducing instruction-based reasoning. The goal of Stage 1 is to establish a stable multimodal grounding between lunar imagery and descriptive geological language.

Our architecture consists of three components (1) a pretrained vision encoder from pretrained CLIP-ViT-Large-Patch14~\cite{radfordclip} that extracts patch-level image representations (2) a learnable multimodal projection layer that maps visual features into the hidden space of the language model (3) a pretrained autoregressive language model that performs text generation. During Stage 1, the pretrained vision encoder extracts patch-level visual features that encode the local terrain structure. These features are mapped into the language model embedding space through a trainable projection layer, allowing the model to associate planetary surface patterns with a domain-specific language. The vision encoder remains frozen throughout this stage in order to preserve its pretrained visual representation capacity.
The base language model parameters are also kept frozen. However, lightweight low-rank adaptation (LoRA)  modules are inserted into the transformer layers and optimized jointly with the multimodal projection layer.

To establish domain-specific grounding, we train Stage 1 on 76k lunar image-caption pairs constructed from our curated planetary dataset. Each sample consists of a geo-referenced lunar surface image tile, along with a structured scientific caption describing terrain morphology, crater density, degradation state, slope distribution, and surface texture where applicable.
Each image-caption pair is converted into an instruction-style format in which the model is prompted to describe the given image. The target output is the original structured caption. This formulation encourages the model to associate planetary surface features with domain language patterns and scientific terminology.

Training is performed using a causal language modeling objective. 
Given an input image $I$ and caption tokens 
$y = \{y_1, \ldots, y_T\}$, 
we minimize the negative log-likelihood:

\begin{equation}
\mathcal{L} = - \sum_{t=1}^{T} 
\log P\!\left(y_t \mid y_{<t}, I \right).
\label{eq:stage1_loss}
\end{equation}

Here, $y_{<t}$ denotes all previously generated tokens and 
$I$ represents the encoded visual features projected into the language embedding space. 
The probability distribution $P(\cdot)$ is modeled autoregressively by the language model, which is conditioned on both the visual representation and the prior tokens.

This stage can be interpreted as aligning visual representations of lunar surface structures with existing language model embeddings in the planetary science domain. At the end of this phase, the model can generate coherent and geologically consistent descriptions of lunar surface structures. However, it is not yet optimized for complex reasoning or conversational interaction. Those capabilities are introduced in Stage~2 through instruction tuning.

\subsubsection{Instruction Tuning Stage}
After alignment with the domain data, we fine-tune the model to operate as a planetary conversational assistant capable of answering scientific questions and performing reasoning over multimodal lunar inputs.

To activate instruction-following behavior, we train on our planetary instruction dataset containing more than 81k question-answer turns. Multiple instructions are associated with each image, allowing the model to learn different reasoning patterns grounded in the same visual context. The instructions span several categories, including surface properties, spatial relationships between geological features, geomorphological characterization, and interpretations of formation processes and surface evolution. This multi-turn supervision trains the model towards scientific reasoning.

Stage~2 is trained using the same causal language modeling objective as defined in \cref{eq:stage1_loss}. However, the supervision shifts from global caption generation to instruction-conditioned question answering. The model is trained to produce targeted, reasoning-based responses grounded in the same visual representations. The loss is applied only to the assistant response tokens, whereas the instruction tokens are treated as a conditioning context.

Stage~2 is initialized from the low-rank adaptation weights obtained after Stage~1. The LoRA modules injected into the transformer layers remain trainable, and the multimodal projection layer continues to update. The base language model parameters remain frozen. This allows the model to refine the domain adaptation while maintaining the stability of the pretrained backbone.

The multimodal projection layer continues to learn a mapping from the vision encoder feature dimension to the hidden dimension of the language model, while the LoRA modules refine internal attention representations, allowing the model to better integrate planetary terminology, geological structure descriptions, and cross-modal reasoning grounded in lunar imagery.

The two-stage design separates the foundational multimodal grounding from the deeper adaptation to reasoning. This separation improves training stability and allows visual alignment to converge before introducing more complex reasoning objectives.

\begin{table*}[t]
\centering
\caption{Performance Ratios Relative to GPT and Gemini Judges (Score / Judge Score)}
\label{tab:category_comparison}
\footnotesize
\begin{tabular}{lcccccccc}
\toprule
& \multicolumn{4}{c}{GPT Judge} & \multicolumn{4}{c}{Gemini Judge} \\
\cmidrule(lr){2-5} \cmidrule(lr){6-9}
Model &\textbf{ \textcolor{blue!40}{Detailed}} & \textbf{ \textcolor{purple!40}{Conversation}} & \textbf{ \textcolor{orange!40}{Reasoning} } & \textbf{Overall} & \textbf{ \textcolor{blue!40}{Detailed}} & \textbf{ \textcolor{purple!40}{Conversation}} & \textbf{ \textcolor{orange!40}{Reasoning} } & \textbf{Overall} \\
\midrule
LLaVA-LE Stage2 & \textbf{0.952} & \textbf{0.767} & \textbf{1.006} & \textbf{0.925} & \textbf{0.892} & \textbf{0.628} & \textbf{1.134} & \textbf{0.917} \\
LLaVA-LE Stage1 & 0.433 & 0.419 & 0.480 & 0.450 & 0.367 & 0.362 & 0.536 & 0.436 \\
Base LLaVA      & 0.319 & 0.314 & 0.344 & 0.329 & 0.220 & 0.205 & 0.246 & 0.227 \\
\bottomrule
\end{tabular}
\end{table*}
\section{Experiments}

We evaluate LLaVA-LE in a controlled setting designed to assess multimodal instruction-following and complex reasoning. Our evaluation focuses on performance under open-ended planetary visual interpretation.

\subsection{Evaluation Benchmark}
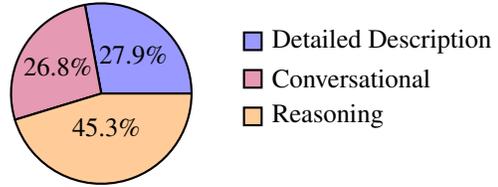
\begin{figure}[t]
    \centering
    \begin{tikzpicture}
        \pie[
            color={blue!40, purple!40, orange!40},
            text=legend,
            radius=1.2 
        ]{
            27.9/Detailed Description,
            26.8/Conversational,
            45.3/Reasoning
        }
    \end{tikzpicture}
    \caption{\textbf{Category distribution of the evaluation set}: Reasoning category is intentionally emphasized in the evaluation set because they pose a harder challenge beyond general visual description.}
    \label{fig:eval_distribution}
\end{figure}

To evaluate domain adaptation and reasoning ability, we constructed a held-out evaluation benchmark of 50 lunar patches from the LUCID stage 1 dataset. Using a language-only GPT-5.1, we generated 190 questions covering the categories shown in Figure~\ref{fig:eval_distribution}. The language model is instructed to generate responses solely based on the scientific caption. This evaluation set was excluded from both stages of training LLaVA-LE.

As depicted in Figure~\ref{fig:eval_distribution}, the reasoning category is intentionally emphasized because it requires interpretation of crater morphology, degradation state, slope distribution, feature relationships, and inferred geological processes. This category represents the most challenging subset and is designed to stress-test multimodal reasoning capability. These questions assess the model’s ability to move beyond visual description and perform scientific interpretation.

\subsection{Evaluation Protocol}

To assess the impact of domain alignment and instruction tuning, we adopt an automatic evaluation protocol based on LLM judges, following the evaluation strategy introduced in LLaVA~\cite{liu2023llava} and LLaVA-Med~\cite{li2023llavamed}.

\paragraph{Candidate Model Responses.}
We compare three variants of the model. \textbf{Base LLaVA} is the original LLaVA-v1.5-13B model without any lunar-specific adaptation, trained on general-purpose multimodal instruction data, and not exposed to planetary science content. It serves as a strong general-domain multimodal baseline. \textbf{Concept-Aligned LLaVA-LE} is our Stage 1 model, trained on caption-based concept alignment data without instruction tuning, which learns to align lunar imagery with scientific descriptions. \textbf{Fully Trained LLaVA-LE} incorporates both Stage 1 alignment and Stage 2 instruction tuning on the LUCID VQA dataset, optimizing the model for multimodal scientific dialogue and reasoning. All models receive the raw lunar image and the corresponding question, without access to the scientific caption. Each model generates a free-form textual response based solely on visual content and the question.

\paragraph{Reference Answers.}
To establish a strong textual reference response for open-ended evaluation, we generated answers using GPT-5.1 and Gemini-2.5 from the ground truth captions in the LUCID dataset.

\paragraph{Automatic Judging.}
To improve robustness and reduce single-model bias, we employ two independent judges: ChatGPT~\cite{chatgpt} and Gemini~\cite{comanici2025gemini}. For each question, an LLM judge receives the scientific caption, the reference answer, and the candidate model responses. Importantly, judges do not access the raw image during scoring; instead, evaluation is grounded in the scientific caption, which serves as a structured textual description of visual content. The agreement between a model's response and the reference answer, therefore, indicates successful visual grounding and domain-specific reasoning. Each judge assigns a score from 1 to 10 based on relevance, clarity, and accuracy.

\subsection{Qualitative Analysis}
To complement quantitative LLM-based scoring, we conduct qualitative analysis to better understand model behavior across question categories. We analyze performance of the base LLaVA model, and our stage 1 and 2 models to illustrate differences. Figure~\ref{fig:qualitative_examples} presents example responses for a selected lunar region. 

\paragraph{Category-Level Performance.}
Table~\ref{tab:category_comparison} reports relative performance across detailed, conversational, and reasoning question types. LLaVA-LE Stage 2 achieves an average overall score of 0.921 (averaged across GPT and Gemini judges), representing a \textbf{3.3$\times$ improvement} over Base LLaVA (0.278) and a \textbf{2.1$\times$ improvement} over LLaVA-LE Stage 1 (0.443). These gains are consistent across all categories, with Stage 2 scoring 0.922 on Detailed ($\sim$3.4$\times$ over Base LLaVA), 0.698 on Conversation ($\sim$2.7$\times$), and most notably \textbf{1.070 on Reasoning ($\sim$3.6$\times$)} --- \textit{exceeding the judge's own reference score} --- demonstrating that Stage 2 instruction tuning is particularly effective at unlocking complex reasoning capabilities that neither Base LLaVA nor Stage 1 can approach. The consistent 2$\times$ margin of Stage 2 over Stage 1 across all categories further highlights the critical role of instruction tuning in transforming a capable vision-language backbone into a model that can engage in detailed, conversational, and reasoning VQA.

To ensure that the inferred subsurface interpretations were scientifically plausible, we manually cross-checked representative examples against independently available gravity and slope data for the corresponding regions. Specifically, we consulted Bouguer gravity anomaly maps derived from the GRAIL mission~\cite{zuber2013grail, konopliv2014high} and terrain slope products computed from LOLA elevation data~\cite{smith2010lola}. This inspection supports that the model's geological reasoning aligns with established lunar geophysical observations.


\section{Conclusions}

Recent advances in vision–language models have enabled powerful multimodal reasoning, yet their application to planetary science remains limited by the lack of large-scale datasets that pair planetary imagery with scientifically meaningful descriptions. In this work, we address this gap by introducing LUCID, a large-scale multimodal lunar dataset consisting of \textbf{96k} high-resolution panchromatic images paired with scientifically grounded captions, along with \textbf{LUCID-VQA}, a complementary benchmark containing \textbf{81k} question–answer pairs. Unlike prior efforts that rely on synthetic or limited datasets, LUCID is curated from real lunar mission products and incorporates geophysical slope and gravity information to ensure captions are grounded in geological observations. Leveraging this dataset, we develop \textbf{LLaVA-LE}, a domain-adapted vision–language model trained through a two-stage curriculum for domain concept alignment and instruction-tuned visual question answering. Experimental results show that \textbf{LLaVA-LE} substantially outperforms the base LLaVA model, achieving a \textbf{3.3× overall} improvement and strong gains in reasoning performance under both GPT and Gemini evaluation frameworks. These findings highlight the importance of domain-specific multimodal supervision for enabling vision–language models in scientific domains. By releasing the dataset, benchmarks, and model weights, we aim to facilitate future research at the intersection of computer vision, multimodal learning, and planetary science.

\clearpage
{
    \small
    \bibliographystyle{ieeenat_fullname}
    \bibliography{main}
}


\end{document}